\documentclass[3p,twocolumn]{elsarticle}
\usepackage{graphicx} 
\usepackage{amsmath}
\DeclareMathOperator*{\argmin}{arg\,min} 

\usepackage{amssymb}
\usepackage[pagebackref=true,breaklinks=true,colorlinks,bookmarks=false]{hyperref}
\usepackage{subcaption}
\usepackage{booktabs}
\usepackage{multirow}
\usepackage{graphicx}
\usepackage{lscape}
\usepackage[acronym,nolist]{glossaries}
\glsdisablehyper 
\newacronym{iot}{IoT}{Internet-of-Things}
\newacronym{hpe}{HPE}{Human Pose Estimation}
\newacronym{mocap}{mocap}{motion capture system}
\newacronym{ml}{ML}{marker-less}
\newacronym{mb}{MB}{marker-based}
\newacronym{nn}{NN}{Neural Network}
\newacronym{ddpm}{DDPM}{Denoising Diffusion Probabilistic Model}
\newacronym{ddim}{DDIM}{Denoising Diffusion Implicit Model}
\newacronym{dm}{DM}{diffusion model}
\newacronym{hri}{HRI}{Human-Robot Interaction}
\newacronym{dl}{DL}{Deep Learning}
\newacronym{dhcf}{DHCF}{dynamic hybrid consensus filter}
\newacronym{bsm}{BSM}{Biomechanical Skeleton Model}
\newacronym{mot}{MOT}{Multi-Object Tracking}

\newcommand*{\draft}{}
\ifdefined\draft
    \usepackage{comment}
    \usepackage[textsize=footnotesize]{todonotes}
    \DeclareRobustCommand{\enrico}[1]
    {{\todo[color=green!40,inline]{Enrico: #1}}} 

    \DeclareRobustCommand{\hojin}[1]
    {{\todo[color=orange!40,inline]{Ho Jin: #1}}} 
    
    \DeclareRobustCommand{\nadia}[1]
    {{\todo[color=yellow!40,inline]{Nadia: #1}}} 

    \DeclareRobustCommand{\nicola}[1]
    {{\todo[color=red!40,inline]{Nicola: #1}}}

    \usepackage[nopar=false]{lipsum}
    \DeclareRobustCommand{\fake}[1]{\color{gray!40!}{\lipsum[][#1]}\color{black}}

\else 
\usepackage{pbalance} 
\newcommand{\enrico}[1]{}
\newcommand{\nicola}[1]{}
\newcommand{\nadia}[1]{}
\newcommand{\hojin}[1]{}
\newcommand{\fake}[1]{}

\fi

\journal{Information Fusion}

\begin{document}
\begin{frontmatter}
\title{COMETH: Convex Optimization for Multiview\\Estimation and Tracking of Humans}
\author{Enrico Martini$^{a,b}$, Ho Jin Choi$^b$, Nadia Figueroa$^b$, Nicola Bombieri$^a$}

\affiliation{organization={Department of Engineering for Innovation Medicine},
            addressline={University of Verona}, 
            city={Verona},
            state={Italy}
            }
            
\affiliation{organization={GRASP Laboratory, Department of Mechanical Engineering and Applied Mechanics},
            addressline={University of Pennsylvania}, 
            city={Philadelphia},
            state={Pennsylvania},
            country={USA}
            }
            
\begin{abstract}
In the era of Industry 5.0, monitoring human activity is essential for ensuring both ergonomic safety and overall well-being. While multi-camera centralized setups improve pose estimation accuracy, they often suffer from high computational costs and bandwidth requirements, limiting scalability and real-time applicability. Distributing processing across edge devices can reduce network bandwidth and computational load. On the other hand, the constrained resources of edge devices lead to accuracy degradation, and the distribution of computation leads to temporal and spatial inconsistencies. We address this challenge by proposing COMETH (Convex Optimization for Multiview Estimation and Tracking of Humans), a lightweight algorithm for real-time multi-view human pose fusion that relies on three concepts: it integrates kinematic and biomechanical constraints to increase the joint positioning accuracy; it employs convex optimization-based inverse kinematics for spatial fusion; and it implements a state observer to improve temporal consistency. We evaluate COMETH on both public and industrial datasets, where it outperforms state-of-the-art methods in localization, detection, and tracking accuracy. The proposed fusion pipeline enables accurate and scalable human motion tracking, making it well-suited for industrial and safety-critical applications. The code is publicly available at~\url{https://github.com/PARCO-LAB/COMETH}.
\end{abstract}

\end{frontmatter}

\section{Introduction}\label{SEC:INTRO}
The emergence of Industry 5.0 has accelerated the integration of non-contact IoT (Internet of Things) devices in industrial environments, enabling smarter, safer, and more efficient workplaces~\cite{Lupion2024}. Human pose tracking is crucial in this transformation, enhancing ergonomic assessments, safety monitoring, and human-robot collaboration~\cite{geretti2022process}. By accurately analyzing workers’ movements, industries can optimize workflows, prevent injuries, and ensure compliance with safety regulations~\cite{rathor2022detailed}. 


Achieving reliable 3D human pose tracking presents significant technical challenges, particularly regarding accuracy, computational efficiency, and scalability~\cite{Bazo2020}.
Traditional multi-camera centralized setups reach high-accuracy 3D human pose estimation (HPE) by fusing information from multiple viewpoints. However, these methods are computationally expensive and require high-bandwidth network transmission, limiting their real-time applicability and scalability in large industrial environments. An alternative approach is edge computing, by which the overall processing is distributed across multiple devices to reduce network congestion and computational load~\cite{zhang2022information}. 
While this strategy alleviates processing and network limitations, edge devices' limited computational power and memory capacity often reduce model accuracy compared to more powerful centralized systems. Achieving a balance between efficiency and precision remains a key challenge in deploying scalable distributed HPE systems~\cite{Xu2018}.

A variety of pose fusion algorithms have been developed to effectively tackle these challenges, integrating multiple sources of pose data to enhance accuracy and robustness. Carraro et al.~\cite{Carraro2019} proposed a fusion algorithm using multiple Unscented Kalman Filters~\cite{wan2000unscented}, each tracking a different joint for every person in the scene. The system uses the 2D skeletons from OpenPose~\cite{Cao2019} and projects the result in 3D using depth data. The system leverages a data association algorithm to match new detections with existing tracks, handle new tracks, or remove old ones. 
Boldo et al.~\cite{Boldo2024} proposed BeFine, a distributed platform designed to perform \gls{hpe} in industrial settings. Their fusion module uses clustering and temporal association algorithms to provide a unique 3D skeleton for each subject in the scene. While these methods have been widely used, they do not provide guarantees for the accuracy of the fused human poses, as they are not built upon a human model.

We address this challenge by proposing COMETH (Convex Optimization for Multiview Estimation and Tracking of Humans), a novel and lightweight algorithm for multi-view human pose fusion for real-time, resource-constrained environments. COMETH is built upon three core concepts to enhance human pose estimation and tracking accuracy and consistency. First, it incorporates kinematic and biomechanical constraints to ensure anatomically plausible and physically consistent joint positioning, even in noisy or partial observations targeting low-power devices. Second, it leverages convex quadratic programming inverse kinematics (QPIK)  to fuse pose estimates from multiple camera perspectives into a unified skeletal representation. Third, it implements a state observer, improving temporal stability and reducing jitter and drift across frames.

We evaluate our approach on the publicly available CMU Panoptic dataset~\cite{Joo_2017_TPAMI} and real-world scenarios, demonstrating improved tracking accuracy and efficiency compared to state-of-the-art methods. Our findings suggest that COMETH is well-suited for real-time applications, including industrial monitoring and interactive systems, where accurate and efficient 3D human pose tracking is essential.

The article is organized as follows. Section~\ref{SEC:SOA} presents an analysis of the related work. Section~\ref{SEC:COMETH} presents COMETH. Section~\ref{SEC:EXPERIMENTS} presents the experimental settings. Section~\ref{SEC:RESULTS} reports the experimental results, while Section~\ref{SEC:CONCLUSION} discusses the conclusions and future directions.


\section{Background}\label{SEC:SOA}
\subsection{3D Human Pose Tracking}

\Gls{mb} motion capture systems are widely employed in motion analysis for research and commercial use~\cite{vicon}. They detect physical markers attached to the subject's body using specialized cameras. Although highly accurate, these systems have significant limitations, including high costs and the need for controlled environments. \Gls{ml} motion capture systems overcome many challenges by being non-intrusive and operating in diverse environments without specialized hardware~\cite{desmarais2021review}. \Gls{hpe} is a computer vision-based \gls{ml} technology that localizes the spatial positions of anatomical keypoints (e.g., elbows, knees, and wrists) of the human body from images or videos~\cite{sarafianos20163d}. Early approaches employed background subtraction and edge detection~\cite{moeslund2006survey}, while state-of-the-art approaches predominantly utilize \gls{dl} approaches~\cite{zheng2023deep}. 

\gls{hpe} has limitations, including external occlusions and self-occluded poses, that can result in incorrect or missing key points~\cite {martini2023denoising}. Several methodologies utilize multi-camera setups to overcome limitations and cover a broader space. Generally, each camera captures input frames and transmits them to a centralized processing module that estimates the 3D pose representation of the subjects~\cite{ zhang2025multi}. 
Although multi-view \gls{hpe} mitigate some challenges of single-view methods, they have some limitations. They are resource-intensive, requiring substantial computational power to process and triangulate all the incoming streaming cameras in real-time. Additionally, network latency and data transfer bottlenecks can further affect the applicability and scalability of those solutions. 


\subsection{Digital Human Representation}
Digital human modeling is critical for \gls{hpe} as it provides a structured representation of the human body. These models enable more precise estimations of joint locations and body poses, correcting the keypoints detected through \gls{hpe}.

Kinematic models represent the human body as a series of linked segments and joints, usually following the Denavit-Hartenberg convention~\cite{denavit1955kinematic}. This method is widely used in human motion analysis and humanoid robots~\cite{Koptev2021} as it constrains the 3D joints' position in space at a fixed distance from each other in correspondence with the bones. Xu et al.~\cite{Xu2018} modeled the human body as a combination of joints and limbs to limit the drift errors of wearable sensors. They used a geometrical constrained method to improve human motion, describing the forward kinematics structure with the Denavit–Hartenberg convention.

\textit{Surface models} use deformable meshes based on shape and pose parameters to represent the body. The most popular surface body model is \textit{Skinned Multi-Person Linear Model (SMPL)}~\cite{SMPL:2015}. It features a fixed topology of 6,890 vertices and 23 joints, enabling realistic animation and deformation. Owing to its efficacy and adaptability, SMPL is widely employed in computer vision, graphics, and applications such as motion capture, virtual reality, and human-centric artificial intelligence. However, it requires accurate parameter estimation, posing challenging computational costs in noisy scenarios. Furthermore, it fails to represent body structure accurately because it treats each joint as a ball joint with three angular degrees of freedom.

Biomechanical models are effective representations for simulating and analyzing human movement, dynamics, and forces~\cite{delp2007opensim}. Werling et al.~\cite{Werling2023} presented \textit{AddBiomechanics}, a method to extract and standardize human movement dynamics. Given a motion capture dataset, the tool scales the body segments of a musculoskeletal model, registers marker locations, and computes body segment kinematics. The accuracy of these approaches heavily depends on the quality of the input data from the \gls{hpe}, which may contain noise, occlusions, or scaling ambiguities, resulting in less reliable biomechanical predictions.

Hybrid models combine surface and biomechanical models, providing visual and biomechanical accuracy. Keller et al.~\cite{keller2023skel} proposed \textit{SKEL}, a combination of the SMPL body model with a biomechanical accurate skeleton. They fit the \gls{bsm} inside SMPL meshes of the AMASS dataset~\cite{mahmood2019amass} and then trained a regressor from SMPL mesh vertices to the optimized joint locations, leading to a more biomechanically realistic representation of bone angles.

\subsection{Inverse Kinematics}
Inverse kinematics (IK) is a mathematical technique determining the joint-space configuration required for a manipulator to reach a desired end position~\cite{siciliano2008springer}. Unlike forward kinematics (FK), which computes the end-effector's position from given joint angles, IK works in reverse, solving for joint configurations based on target positions. For highly redundant manipulators, IK often involves optimization techniques such as gradient descent or Jacobian pseudoinverse algorithms. Convex optimization, a field of mathematical optimization, focuses on minimizing convex functions within convex sets. A specific type of convex optimization, quadratic programming (QP), seeks to minimize a quadratic function while satisfying linear constraints~\cite{boyd2004convex}. In convex QP, any local minimum is also the global minimum.
Quadratic Programming Inverse Kinematics (QPIK) is an advanced IK method that formulates the IK problem as a QP problem. It is commonly used in robotics to enforce additional constraints, such as joint limits, balance, and collision avoidance~\cite{Koptev2021}.


\section{COMETH}\label{SEC:COMETH}
\begin{figure*}[t]
    \centering
    \includegraphics[width=\linewidth]{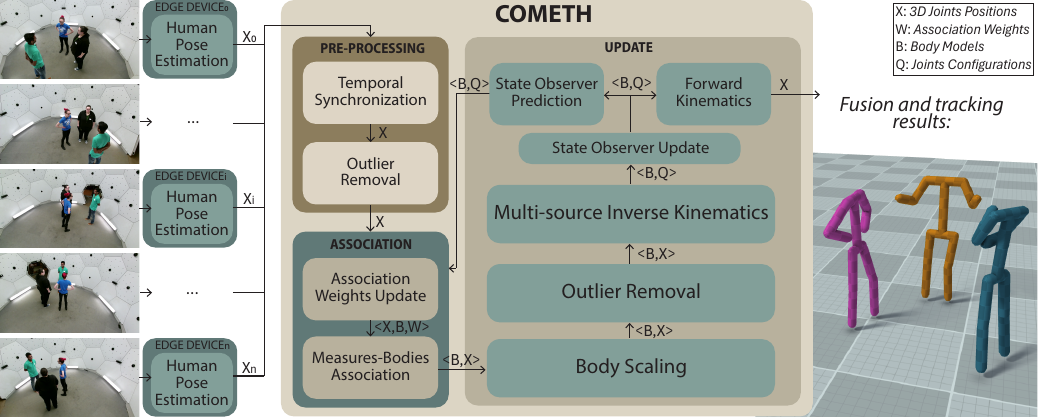}
    \caption{Overview of the multi-camera fusion pipeline. Edge devices independently perform HPE from multiple camera views. The aggregator applies temporal synchronization and outlier removal before associating detected poses $X$ with previous body models $B$ through the target association module. In the model's update stage, a state observer refines motion consistency, while multi-source inverse kinematics obtains the current joint configuration, correcting errors via bone scaling and additional outlier removal. Finally, forward kinematics reconstructs the accurate poses in 3D space. The right side illustrates a representation of the pipeline results.}
    \label{fig:overview}
\end{figure*}

\textit{COMETH (Convex Optimization for Multiview Estimation and Tracking of Humans)} is an aggregation algorithm that fuses information from multiple edge devices. At each time frame, the devices send sets of keypoints representing the 3D coordinates of human joints if they detect people in the scene. The aggregator has two steps: first, it synchronizes incoming detections in space and time and pairs them to existing bodies; second, it updates the positions of these bodies based on the associated measurements.  Figure \ref{fig:overview} shows an overview of the proposed fusion method, with each block described in detail over the following sections.

\subsection{Pre-processing}\label{sec:pre-processing}
At time $t$, each edge device may send a set of measurements $X$ over the network if a person is captured, where $|X|$ is the number of people detected by the \gls{hpe}. Each measurement $x \in X$ comprises a set of labeled 3D points representing human joints, called keypoints. Since different edge devices with varying computing capabilities may be deployed in the scene, the information may arrive with various latencies and frequencies. When COMETH receives new measurements, they are stored in an ordered queue. At each iteration of the aggregator, measurements are removed from the queue if considered too old, using a time duration threshold $\varDelta$. Given the aggregator timestamp $t_a$, a set of measurements $X_n$ captured by the device $n$ at time $t_{X_n}$, if $t_a-t_X<\varDelta$ then $X$ is considered synchronized and ready to be fused. Otherwise, it is discarded. 

The pre-processing function then analyzes all the selected measurements, removing outliers and false detections caused by the \gls{hpe} or the depth sensor. This function eliminates keypoints that are too distant from the camera. This step is essential because even the most advanced depth sensors exhibit inaccuracies at greater distances from the target.

\subsection{Association}
All measurements with at least $k$ valid keypoints are fed into the association weights function to be associated with their body models.  

Each body model $b \in B$ is derived from BSM~\cite{keller2023skel}, a digital human representation with 49 degrees of freedom (DOFs). The first 6 DOFs represent the absolute position and orientation in 3D space, while the remaining 43 correspond to anatomically accurate body joints connecting 24 rigid bone groups. 

We use the Hungarian algorithm for each edge device $i\in[0,n)$ to find an assignment between the body models $B$ and the measurements $X_i$.
Given a $|B|\times|X_i|$ cost matrix $W_i \in \mathbb{N}^+$, the weight matrix is computed as follows:
\begin{align}
W_i[k,j] = \kappa_2{(||X_i[k]-\text{FK}(B_j)||_2)}
\end{align}
where $\text{FK}(\cdot)$ is the forward kinematics of the body model, and $\kappa_2(\cdot)$ is a function that selects the second smallest element from a set. We select the second closest value between the two sets of 3D points based on their Euclidean distances, as the smallest distance could be an outlier.

Then, for each body model $b \in B$, we associate a set of measurements $X_b$ that might belong to the subject represented by $b$. Given that there may be at most one detection per subject for each edge device $i\in[0,n)$, at the end of the association process, each $b \in B$ may have a set of measurements $X_b$, where $|X_b| \in [0,n)$. After the association process, we create a new body model for each unassociated set of measurements.

\subsection{Update}
During this phase, COMETH updates each body's information based on the associated measurements. We update the bones' scale for each body, removing the incompatible keypoints. Next, we retrieve all joint displacement values through QPIK. A state observer then filters the QPIK results for temporally smoothed values. Finally, forward kinematics accurately reconstructs the poses in 3D space.

\subsubsection{Body Scaling and Outlier Removal}\label{ss:bone_scaling}
\begin{figure}
    \centering
    \includegraphics[width=\linewidth]{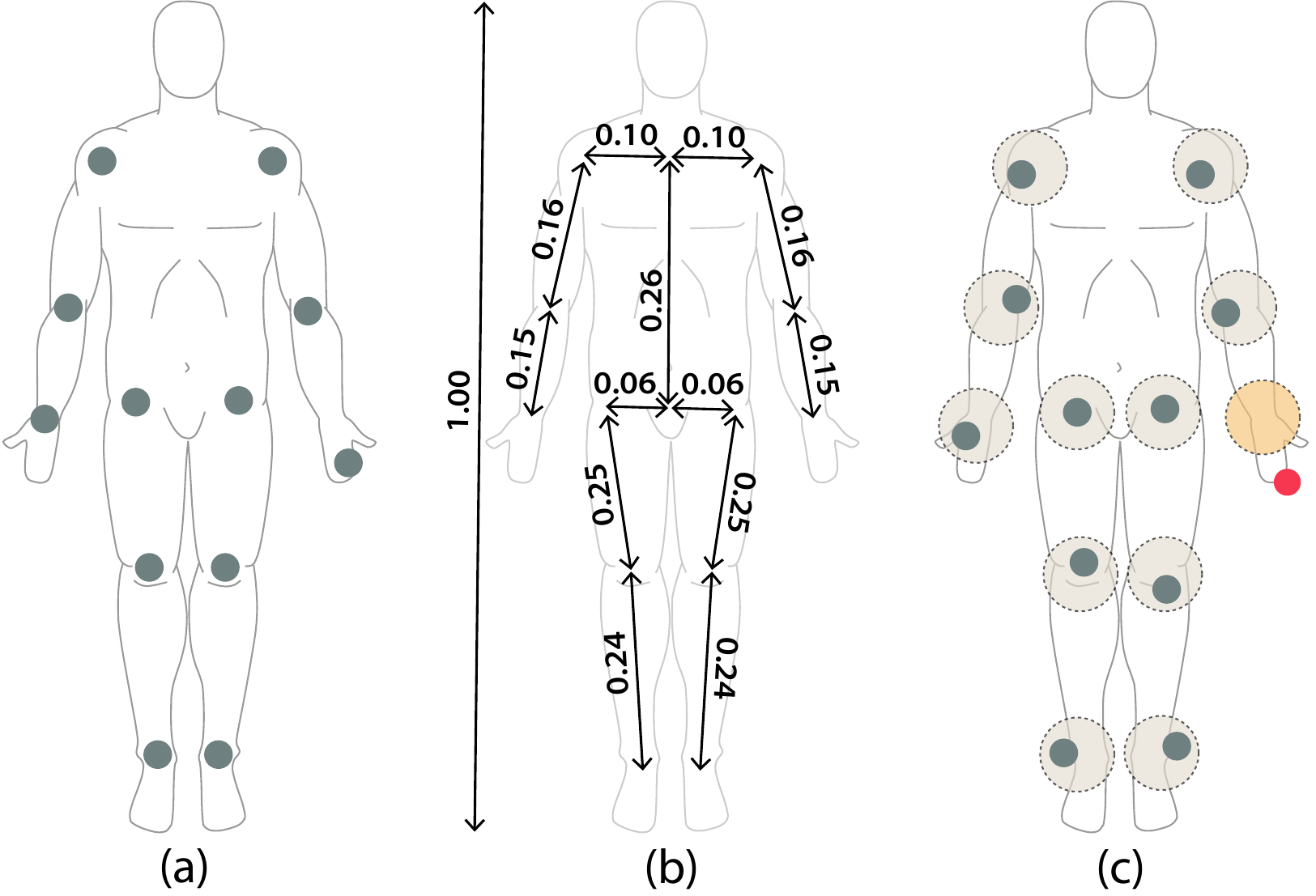}
    \caption{Graphical representation of the body scaling and outlier removal phase: the input measurements (a) are used to estimate the height of the subject following fixed proportions (b). Based on these estimates, keypoints that fall outside acceptable distance ranges are identified and removed as outliers (c).}
    \label{fig:scaling}
\end{figure}

For each body model $b\in B$, we update the scale of the bones.
Scaling the bones of the body model to fit input data is a crucial step to ensure that the model accurately reflects the subject's specific dimensions and geometry. Estimating body scale from clean input using an optimization algorithm is straightforward; however, handling partial or noisy data presents challenges. First, we begin by estimating the subject's height. Since segment lengths have a well-defined relationship with overall human height~\cite{drillis1964body, winter2009biomechanics}, we use them to approximate the subject's height based on the inter-joint distances, as in~\cite{martini2024robust}. Each distance between two physically connected joints is used to estimate height, and we compute a reasonable height by averaging all possible estimations. The average scaling factor, $s_{AVG}$, is derived from this value.
To account for individual anatomical variations, we apply different scaling factors to each bone rather than a uniform scale. For instance, two people of the same height may have significantly different limb lengths, affecting joint positioning and movement patterns. Thus, for each distance between adjacent keypoints (e.g., knee and ankle), we compute the corresponding body part's scaling factor and clip it within $\pm5\%$ of $s_{AVG}$. For body parts that cannot be directly inferred from keypoints, we use $s_{AVG}$. Then, we remove outliers from the measurements $X_b$, excluding all keypoints that form bones with lengths incompatible with the bones of $b$. 
We store previous scaling factors and compute their average over time to ensure consistency in representing the same subject across frames. Figure~\ref{fig:scaling} depicts the body scaling process and the removal of the outliers.

\subsubsection{Multi-Source Inverse Kinematics}\label{sec:msqpik}

In traditional robotics IK problems, the objective is to determine a joint configuration that positions the end effector at a desired target location with a specific orientation. To determine the joint configuration of a human model from HPE data, each joint $i$ has its own positional target keypoint $kp_i$. Given a set of 3D points $x_T \in \mathbb{R}^{3|kp|}$, where $|kp|$ is the number of keypoints detected, we propose the following iterative QP formulation to solve the IK:
\begin{align}
\argmin_{\dot{q},\delta}{~\dot{q}^T \Lambda \dot{q} + \delta^T D \delta}\label{eq:qpik_obj}
\end{align}
\vspace{-15pt}
\begin{align}
\text{s.t.}~
\begin{cases}
x + J(q)\dot{q} = x_T + \delta\\
-\gamma(q-q_L) \le \dot{q} \le \gamma(q_U - q)\\
\dot{q}_L - \dot{q}_0 \le \dot{q} \le \dot{q}_U - \dot{q}_0\label{eq:qpik_constraints}
\end{cases}
\end{align}
where $q\in \mathbb{R}^{49}$ is the angular displacement, $\dot{q}\in \mathbb{R}^{49}$ is the angular velocity, $\delta \in \mathbb{R}^{3|kp|}$ is a slack variable, $D\in \mathbb{R}^{3|kp|\times 3|kp|}_{++}$ is a symmetric positive definite matrix that sets weights for the slack variable $\delta$, and $\Lambda\in \mathbb{R}^{49\times 49}_{++}$ is a symmetric positive definite matrix equivalent to the damping term in least-squares IK methods.
Equation~\ref{eq:qpik_constraints} describes the three constraints to obtain a realistic configuration for the human model. The first one represents the linear approximation of the relationship between the Cartesian position and the angular displacement of each joint. The current position $x\in \mathbb{R}^{3|kp|}$, when updated by the motion determined by the product of the Jacobian matrix $J\in \mathbb{R}^{3|kp|\times 49}$ and joint velocities $\dot{q}\in \mathbb{R}^{49}$, should match the target position $x_T$ with a small error $\delta$.

The second constraint defines dynamic limitation on joint velocities to keep the joints within their physical range. It guarantees that the speed of a joint is appropriately reduced as it approaches its limits. A limited Range of Motion (ROM) ensures that our model accurately reflects human anatomical constraints. We define the lower limits $q_L\in \mathbb{R}^{49}$ and upper limits $q_U\in \mathbb{R}^{49}$ for each joint position based on normative data from the biomechanics literature. For instance, unlike BSM, COMETH restricts the elbow flexion to a range between -11° and 154°, following the normative data presented in~\cite{zwerus2019normative}. This range accounts for physiological variations while preventing unrealistic joint movements and reducing the search space for the inverse kinematics solver. The scaling factor $\gamma \in (0,1]$ controls how aggressively the joint approaches its limits, acting as a barrier function \cite{Ames2019ControlBF}.

In addition to joint positions, we impose velocity constraints. Joint velocity limitations arise from several anatomical factors, including muscle contraction speed and tendon elasticity. These constraints vary across joints; for example, the wrist can execute rapid small movements, while the knee moves more slowly due to greater mass and structural constraints. Velocity limits vary across individuals, influenced by flexibility, muscle strength, and training history. Since the biomechanics literature on every joint velocity is limited, we regress them from the BioAMASS dataset~\cite{keller2023skel}. This comprehensive motion capture dataset extends the AMASS dataset~\cite{mahmood2019amass}, focusing on biomechanical applications. It includes high-quality kinematic and dynamic recordings of human motion from individuals with diverse anthropometric characteristics, including variations in age, gender, and physical condition. BioAMASS comes with BSM joint positions as pseudo-ground truth, so we apply the same scaling from BSM to each subject in COMETH. To align the COMETH body model with the ground truth, we employ an inverse kinematics solver based on the Levenberg-Marquardt algorithm~\cite{levenberg1944method}. For each sequence in the dataset, we extract instantaneous joint velocities. The lower bound of angular velocity $\dot{q}_L\in \mathbb{R}^{49}$ is determined using the 5th percentile, while the upper bound $\dot{q}_U\in \mathbb{R}^{49}$ is set using the 95th percentile. The third constraint sets dynamic bounds on the change in the final joint velocity, represented as $\dot{q}_0 + \dot{q}$. Here, $\dot{q}_0$ denotes the aggregated velocity over a single timestep, accounting for multiple iterations of the QPIK solver. Considering the current state, this constraint ensures that the resulting velocity stays within the limits. Both ROM and velocity constraints are reported for each joint angle in Table~\ref{tab:position-velocity-limits}.

\paragraph{Multi-source QPIK}\label{ss:ms-qpik}
When solving the IK with multiple targets, there may be conflicting constraints, requiring strategies such as constraint relaxation~\cite{boyd2004convex}. To determine the joint configuration of a human model from multiple HPE devices, each joint $i$ may have one or more target positions.
We formulate a QPIK that considers mutiple sets of 3D points $x_{Ti}$ coming from $n$ sources as follows:
\begin{align}
\argmin_{\dot{q},\delta_0,\dots,\delta_n}{~\dot{q}^T \Lambda \dot{q} + \sum_{i=1}^{n}\delta_i^T D \delta_i}\label{eq:mqpik_obj}
\end{align}
\vspace{-10pt}
\begin{align}
\text{s.t.}~
\begin{cases}
 \forall i \in [1,n] : x_i + J_i(q) \dot{q} = x_{\text{Ti}} + \delta_i \\
-\gamma(q-q_L) \le \dot{q} \le \gamma(q_U - q)\\
\dot{q}_L - \dot{q}_0 \le \dot{q} \le \dot{q}_U - \dot{q}_0\label{eq:mqpik_constraints}
\end{cases}
\end{align}
Differently from Equation~\ref{eq:qpik_obj}, in~\ref{eq:mqpik_obj} there are $n$ slack variables, one per input source. This formulation minimizes the difference between the final position of each joint and the targets. In Equation~\ref{eq:mqpik_constraints}, we extend the first constraint from Equation~\ref{eq:qpik_constraints} to handle multiple positional targets. The other two constraints remain as in Equation~\ref{eq:qpik_constraints}, since they do not depend on the number of target sources.

\begin{table}[t]
\centering
\caption{Position and velocity limits for each joint angle. Position limits come from the biomechanics literature, while velocity ones are retrieved from the BioAMASS dataset~\cite{keller2023skel}.}
\label{tab:position-velocity-limits}
\resizebox{\linewidth}{!}{
\begin{tabular}{ccccc}
\toprule
\textbf{Joint angle} & \multicolumn{1}{c}{\textbf{\begin{tabular}[c]{@{}c@{}} $q_L$ \\ $[$°$]$\end{tabular}}} & \multicolumn{1}{c}{\textbf{\begin{tabular}[c]{@{}c@{}} $q_U$ \\$[$°$]$\end{tabular}}} & \multicolumn{1}{c}{\textbf{\begin{tabular}[c]{@{}c@{}} $\dot{q}_L$  \\$[$°/s$]$\end{tabular}}} & \multicolumn{1}{c}{\textbf{\begin{tabular}[c]{@{}c@{}} $\dot{q}_U$ \\$[$°/s$]$\end{tabular}}}\\
\midrule
Hip Flexion/Extension           & $-40$             & $140$        & $-1.6$        & $1.9$    \\
Hip Adduction/Abduction         & $-45$             & $45$         & $-0.6$        & $0.5$    \\
Hip Rotation                    & $-45$             & $45$         & $-0.6$        & $0.5$    \\
Knee Flexion/Extension          & $-10$             & $140$        & $-2.0$        & $2.1$    \\
Lumbar Bending                  & $-20$             & $20$         & $-0.5$        & $0.5$    \\
Lumbar Twist                    & $-5$              & $5$          & $-0.3$        & $0.3$    \\
Thorax Bending                  & $-20$             & $20$         & $-0.4$        & $0.4$    \\
Thorax Twist                    & $-5$              & $5$          & $-0.3$        & $0.3$    \\
Shoulder Adduction/Abduction    & $0$               & $150$        & $-0.8$        & $0.8$    \\
Shoulder Rotation               & $-70$             & $90$         & $-0.9$        & $0.9$    \\
Shoulder Flexion/Extension      & $-60$             & $180$        & $-1.4$        & $1.4$    \\
Elbow Flexion/Extension         & $-11$              & $154$        & $-1.4$        & $1.4$    \\
\bottomrule 
\end{tabular}
}
\end{table}


\subsubsection{State Observer}\label{ss:state-observer}
When dealing with multiple streams of targets in the proposed human model, applying QPIK at each time step may lead to inconsistent solutions due to measurement errors. To deal with these limitations, we integrate QPIK into a state observer to ensure stable and realistic motion reconstruction. A state observer helps manage sensor noise, predict unmeasured joint states, and enforce temporal consistency.  We implement a linear Kalman Filter (KF)~\cite{kalman1960new}, a popular state observer for estimating the state of a dynamic linear system in the presence of noise and uncertainty. For each joint $i$, we formulate the state $Q_i \in Q$, the second-order state transition matrix $F$, and the output matrix $H$ as:
\begin{align}
Q_i &= \begin{bmatrix}
q_i\\
\dot{q}_i\\
\ddot{q}_i
\end{bmatrix},
&
F &= \begin{bmatrix}
&1 &d t &\frac{1}{2}d t^2\\
&0 &1 &d t\\
&0 &0 &1
\end{bmatrix},
&
&
H = \begin{bmatrix}
1\\
0\\
0
\end{bmatrix}^T
\end{align}
where $q_i$ is the positional value of joint angle $i$, $\dot{q}_i$ its velocity and $\ddot{q}_i$ its acceleration.

State observers work through two phases: prediction and correction. In the prediction phase, the current state estimates $X$ are projected forward in time, anticipating how the state will evolve. At time $t$, the predicted state $X^-_i[t]$ is obtained by:
\begin{align}
Q^-_i[t] = FQ_i^+[t-1]\label{eq:kf_prediction}
\end{align}
where $Q_i^+[t-1]$ is the state of the joint angle $i$ at time $t-1$.
In this phase, the covariance estimate $P_i$ is also calculated, which propagates uncertainty over time:
\begin{align}
P^-_i[t] = FP^+_i[t-1]F^T+\varSigma
\end{align}
where $\varSigma$ represents the covariance matrix of the process noise. The prediction accuracy forms the basis for subsequent adjustments made during the correction phase, where the values are refined using real-time measurements.

In the correction phase, the filter computes the Kalman gain $K_i[t]$, determining how much the measurement should influence the estimate based on its reliability:
\begin{align}
K_i[t] = P^-_i[t]H^T(HP_i^-[t]H^T+R)^{-1}
\end{align}
where $R$ is the covariance matrix of the measurement noise.
The updated state $Q^+_i[t]$ is then obtained by adjusting the predicted state $Q^-_i[t]$ with the difference between the result of the QPIK $z_i[t]$ and $Q^-_i[t]$:
\begin{align}
Q^+_i[t] = Q^-_i[t] \cdot K_i[t](z_i[t]-HQ^-_i[t])\label{eq:kf_update}
\end{align}
The covariance matrix is then updated to reduce the uncertainty after incorporating the new information:
\begin{align}
P^+_i[t] = (I - K_i[t]H)P^-_i[t]
\end{align}
This method stabilizes the estimated state of each of the 49 joint angles over time.


\section{Experimental Setup}\label{SEC:EXPERIMENTS}
\subsection{Datasets}
We quantitatively validated the proposed fusion method on the CMU Panoptic Dataset~\cite{Joo_2017_TPAMI}. It includes multiple individuals engaged in social interactions, captured with 480 VGA cameras, 30 HD cameras, and 10 RGB-D Kinect v2 sensors. Our experiments used all sequences with valid ground truth and a stream of 5 RGB-D cameras at 30 Hz. We selected the upper ones as they more closely resemble surveillance cameras than those positioned at a lower height. Depending on the sequence, there may be at the same time 2,3, or 7 actors in the scene. 

We also present the qualitative results of our method deployed in the ICE Laboratory~\cite{Cunico2024}, which is a modern, compact, and realistic production line environment. The setup includes 9 RGB cameras and 7 RGB-D cameras. We employed 3 StereoLabs ZED2 RGB-D cameras\footnote{\url{https://www.stereolabs.com/zed-2/ }} at 15 Hz, each paired with an Nvidia Jetson Xavier NX\footnote{\url{https://www.nvidia.com/en-us/autonomous-machines/embedded-systems/jetson-xavier-nx}} running an HPE node. 

\subsection{Evaluation metrics}
We quantitatively evaluated the proposed method using four metrics from the multiple objects tracking (MOT) literature~\cite{Luiten2021}. First, we define a similarity score $\mathcal{S}$ between a predicted skeleton (\textit{pr}) and its ground truth skeleton (\textit{gt}) as:
\begin{align}
    \mathcal{S}(\text{\textit{pr}},\text{\textit{gt}}) = \max\left(0, 1 - \frac{1}{|\text{kp}|} \sum_{i \in\text{kp}} \| \text{\textit{pr}}_i - \text{\textit{gt}}_i \| \right)
\end{align}
where $\text{kp}$ is the common set of detected keypoints.
Following this notation, when $\mathcal{S}$ is 1, the skeletons perfectly align, and when S is 0, there is no overlap between detections~\cite{Luiten2021}. We perform a bijective matching between each \textit{pr} and \textit{gt}, utilizing the Hungarian algorithm~\cite{kuhn1955hungarian}. Given a similarity threshold $\alpha \in (0,1)$, each matched pair is considered true positive ($TP$) if $\mathcal{S}(\text{\textit{pr}},\text{\textit{gt}}) > \alpha$. Any \textit{gt} that is not matched is a false negative ($FN$), while any \textit{pr} that is not matched is a false positive ($FP$).
The detection accuracy score (DetA) is defined as the standard Jaccard index:
\begin{align}
\text{DetA} =  \int_{0}^{1}{\frac{|TP_{\alpha}|}{|TP_{\alpha}|+|FN_{\alpha}|+|FP_{\alpha}|}~d\alpha}
\end{align}
It measures how accurately the tracker detects skeletons in each frame over time.

The localization accuracy score (LocA) is calculated as follows:
\begin{align}
\text{LocA} =  \int_{0}^{1}{\frac{1}{|TP_{\alpha}|}\sum_{c~\in~TP_{\alpha}}{S(c,\text{\textit{gt}})}~d\alpha}
\end{align}
It measures how accurately the predicted keypoints positions match the ground truth locations.

When considering the ID association, each prediction \textit{pr} is considered a True Positive Association (TPA) if \textit{pr} is in the $TP$ set and has the same id as its \textit{gt}. The set of False Negative Associations (FNAs) is the set of \textit{gt} with the same id as \textit{pr} but either wrongly associated or not associated with any \textit{pr}. The set of False Positive Associations (FPAs) is the set of \textit{pr} with the same ID as \textit{gt} but either wrongly associated or not associated with any \textit{gt}. Given a true positive candidate $c\in TP$, we define the association score $\mathcal{A}$ as:
\begin{align}
    \mathcal{A}(c) = \frac{|TPA(c)|}{|TPA(c)|+|FNA(c)|+|FPA(c)|}
\end{align}
where $\mathcal{A}$ measures the alignment between \textit{gt} and \textit{pr} trajectories that are matched at the $TP$ $c$, calculated using the Jaccard index.
The association accuracy score (AssA) is then defined as:
\begin{align}
\text{AssA} =  \int_{0}^{1}{\frac{1}{|TP_{\alpha}|}\sum_{c \in TP_{\alpha}}{\mathcal{A}(c)}~d\alpha}
\end{align}
It measures how accurately a tracking algorithm maintains the correct associations between detections across frames. 

Higher Order Tracking Accuracy (HOTA) is the geometric mean of detection and association scores:
\begin{align}
    \text{HOTA} = \int_{0}^{1}{\sqrt{ \text{DetA}_\alpha\cdot\text{AssA}_\alpha}}~d\alpha
\end{align}
HOTA evaluates how well tracking algorithms perform by measuring their accuracy in localization and association.

\subsection{Compared methods and baselines}
To maximize compatibility with most \gls{hpe} methods, we consider the common 12 keypoints (shoulders, hips, elbows, wrists, knees, and ankles) as input data. 
We use TRTPose~\footnote{\url{https://github.com/NVIDIA-AI-IOT/trt_pose}} as a baseline for evaluating the Panoptic dataset and OpenPose~\cite{Cao2019}  for the real-case study. In both cases, we employ the 2D HPE to extract 2D keypoints representing human body joints from the RGB video stream. Then, we use the depth matrix to determine each keypoint's 3D distance from the camera. To obtain the 3D coordinates of each keypoint, we back-project the 2D keypoints into 3D space. The final output is a set of \textit{3D keypoints} for each video frame, timestamped, representing the body joints of each detected person in the scene. 

We compare the results of our method with two fusion algorithms that take as input 3D skeletons. The first one is \emph{OpenPTrack}, proposed by Carraro et al.~\cite{Carraro2019}, which combines the Hungarian algorithm and multiple Unscented Kalman Filters to associate and track the subjects. 
The second method, called \emph{Befine}, proposed by Boldo et al.~\cite{Boldo2024}, employs spatial clustering and specific temporal association algorithms.

\subsection{Implementation details}
We implement COMETH upon Nimble~\cite{Werling2021}, a differentiable physics engine derived from DART~\cite{Lee2018}, designed explicitly for articulated rigid body simulation.
We set all the parameters once and kept the same values throughout the experimental results. To solve the QPIK formulation, we utilize \verb|cvxpy|~\cite{diamond2016cvxpy}, a domain-specific language for convex optimization problems that generates efficient C codes tailored for the specific formulation of the optimization problem. All experiments are run on a desktop setup (AMD Ryzen 9 7950X, 64GB RAM DDR5, Nvidia RTX 4090), while the HPE runs on multiple Nvidia Jetson Xavier NX that act as edge devices.

We set all COMETH parameters through grid search, keeping the same values throughout the experiments in both case studies. In detail, in the pre-processing phase of the aggregator (Section \ref{sec:pre-processing}), we set the duration timeout $\varDelta$  to 0.07 seconds.
Regarding the multi-source QPIK (Section \ref{sec:msqpik}), we set the maximum number of iterations to 100. Both matrices $\Lambda$ and $D$ are set to identity matrices, and the scaling factor $\gamma$ is set to 1. In the state observer (Section \ref{ss:state-observer}), the covariance matrix $Q$ is set to a diagonal matrix with diagonal elements equal to 0.5, and the covariance matrix $R$ is set to the identity matrix. 

\section{Experimental Results}\label{SEC:RESULTS}
\subsection{Analysis on CMU Panoptic Dataset}
\begin{table}[t]
\centering
\caption{Results on the panoptic dataset.}
\label{tab:my-table}
\resizebox{\columnwidth}{!}{%
\begin{tabular}{clcccc}
\toprule
\multicolumn{1}{l}{\textbf{Cams}} & \textbf{Method} & \multicolumn{1}{c}{\textbf{\begin{tabular}[c]{@{}c@{}}LocA\\ (\%)$\uparrow$\end{tabular}}} & \multicolumn{1}{c}{\textbf{\begin{tabular}[c]{@{}c@{}}DetA\\ (\%)$\uparrow$\end{tabular}}} & \multicolumn{1}{c}{\textbf{\begin{tabular}[c]{@{}c@{}}AssA\\ (\%)$\uparrow$\end{tabular}}} & \multicolumn{1}{c}{\textbf{\begin{tabular}[c]{@{}c@{}}HOTA\\ (\%)$\uparrow$\end{tabular}}} \\
 \midrule
 & 3D HPE & 76.9 & 56.7 & - & - \\
\midrule
\multirow{3}{*}{1} & OpenPTrack~\cite{Carraro2019} & 74.9 & 40.7 & 43.1 & 40.6 \\
 & Befine~\cite{Boldo2024} & 76.9 & 56.6 & 26.4 & 34.9 \\
 & \textbf{COMETH} & \textbf{78.4} & \textbf{58.9} & \textbf{56.1} & \textbf{56.1} \\
\midrule
\multirow{3}{*}{2} & OpenPTrack~\cite{Carraro2019} & 76.8 & 34.1 & 28.6 & 29.6 \\
 & Befine~\cite{Boldo2024} & \textbf{82.9} & 67.0 & 29.5 & 38.8 \\
 & \textbf{COMETH} & 82.0 & \textbf{69.0} & \textbf{75.8} & \textbf{71.8} \\
\midrule
\multirow{3}{*}{3} & OpenPTrack~\cite{Carraro2019} & 77.1 & 25.1 & 22.4 & 22.0 \\
 & Befine~\cite{Boldo2024} & 84.6 & 68.4 & 30.7 & 39.1 \\
 & \textbf{COMETH} & \textbf{85.1} &\textbf{ 72.0} & \textbf{84.1} & \textbf{77.3} \\
\midrule
\multirow{3}{*}{4} & OpenPTrack~\cite{Carraro2019} & 77.0 & 17.4 & 18.9 & 16.3 \\
 & Befine~\cite{Boldo2024} & 85.1 & 66.2 & 31.0 & 38.1 \\
 & \textbf{COMETH} & \textbf{86.7} & \textbf{72.0} & \textbf{87.1} & \textbf{78.5} \\
\midrule
\multirow{3}{*}{5} & OpenPTrack~\cite{Carraro2019} & 76.7 & 11.4 & 18.4 & 12.7 \\
 & Befine~\cite{Boldo2024} & 85.5 & 63.6 & 31.2 & 37.1 \\
 & \textbf{COMETH} & \textbf{87.4} & \textbf{69.3} & \textbf{86.7} & \textbf{76.7} \\
\bottomrule 
\end{tabular}%
}
\end{table}

Table~\ref{tab:my-table} presents the performance comparison of COMETH against the baseline approaches across different numbers of camera views. With a single camera, our approach reaches a LocA of 78.4\%, which increases to 87.4\% when using five cameras. 
For a lower number of cameras, the LocA of COMETH is similar or superior to that of the others.  When using 2 cameras, Befine performs negligibly better (0.9\%), because of its model-free clustering method that preserves better the initial detections of the HPE. However, compared to the other methods, the robustness of COMETH in correctly localizing human poses increases consistently as the number of cameras increases. 
Similarly, our approach shows superior DetA performance, starting at 58.9\% for a single camera and reaching 69.3\% for five cameras. This indicates a higher reliability in detecting individuals accurately compared to prior methods. The AssA score of our method shows significant improvement over competing approaches, particularly as the number of cameras increases. The AssA score rises from 56.1\% with one camera to 86.7\% with five cameras, demonstrating more robust multi-view association capabilities than the other methods.  COMETH consistently achieves the best HOTA score, reaching 76.7\% with five cameras. This metric reflects the overall tracking performance and confirms the superiority of COMETH in maintaining identity consistency over time.

\begin{table*}[t]
\centering
\caption{Localization and tracking accuracy of the fusion method varying number of subjects and cameras.}
\label{tab:people}
\resizebox{.99\textwidth}{!}{%
\begin{tabular}{clcccccccccc}
\toprule
\multicolumn{2}{c}{\textbf{Cameras}} & \multicolumn{2}{c}{\textbf{1}} & \multicolumn{2}{c}{\textbf{2}} & \multicolumn{2}{c}{\textbf{3}} & \multicolumn{2}{c}{\textbf{4}} & \multicolumn{2}{c}{\textbf{5}} \\
\textbf{Subjects} & \multicolumn{1}{c}{\textbf{Method}} & \textbf{\begin{tabular}[c]{@{}c@{}}LocA\\ (\%)$\uparrow$\end{tabular}} & \textbf{\begin{tabular}[c]{@{}c@{}}HOTA\\ (\%)$\uparrow$\end{tabular}} & \textbf{\begin{tabular}[c]{@{}c@{}}LocA\\ (\%)$\uparrow$\end{tabular}} & \textbf{\begin{tabular}[c]{@{}c@{}}HOTA\\ (\%)$\uparrow$\end{tabular}} & \textbf{\begin{tabular}[c]{@{}c@{}}LocA\\ (\%)$\uparrow$\end{tabular}} & \textbf{\begin{tabular}[c]{@{}c@{}}HOTA\\ (\%)$\uparrow$\end{tabular}} & \textbf{\begin{tabular}[c]{@{}c@{}}LocA\\ (\%)$\uparrow$\end{tabular}} & \textbf{\begin{tabular}[c]{@{}c@{}}HOTA\\ (\%)$\uparrow$\end{tabular}} & \textbf{\begin{tabular}[c]{@{}c@{}}LocA\\ (\%)$\uparrow$\end{tabular}} & \textbf{\begin{tabular}[c]{@{}c@{}}HOTA\\ (\%)$\uparrow$\end{tabular}} \\
 \midrule
\multirow{3}{*}{2} & OpenPTrack~\cite{Carraro2019} & 73.4 & 38.0 & 75.8 & 23.8 & 76.8 & 18.6 & 76.9 & 14.9 & 76.6 & 12.6 \\
 & Befine\cite{Boldo2024} & 75.8 & 35.5 & \textbf{82.2} & 38.0 & 83.5 & 38.2 & 83.7 & 37.0 & 83.7 & 36.0 \\
 & \textbf{COMETH} & \textbf{77.4} & \textbf{50.4} & 81.3 & \textbf{72.5} & \textbf{84.5} & \textbf{81.9} & \textbf{86.3} & \textbf{85.5} & \textbf{87.3} & \textbf{83.6} \\
\midrule
\multirow{3}{*}{3} & OpenPTrack~\cite{Carraro2019} & 75.1 & 39.7 & 75.9 & 28.6 & 75.7 & 20.4 & 75.7 & 15.0 & 76.2 & 12.5 \\
 & Befine\cite{Boldo2024} & 76.3 & 22.5 & \textbf{81.7} & 23.7 & 82.7 & 24.1 & 82.9 & 24.2 & 83.6 & 25.1 \\
 & \textbf{COMETH} & \textbf{78.0} & \textbf{56.4} & 81.1 & \textbf{68.9} & \textbf{84.1} & \textbf{74.7} & \textbf{85.6} & \textbf{76.8} & \textbf{86.1} & \textbf{74.8} \\
\midrule
\multirow{3}{*}{7} & OpenPTrack~\cite{Carraro2019} & 78.1 & 43.6 & 79.7 & 37.1 & 80.0 & 27.7 & 80.0 & 21.5 & 79.3 & 17.0 \\
 & Befine\cite{Boldo2024} & 80.2 & 17.5 & 84.3 & 17.2 & 85.7 & 16.5 & 86.1 & 15.9 & 86.2 & 15.5 \\
 & \textbf{COMETH} & \textbf{81.1} & \textbf{58.8} & \textbf{84.5} & \textbf{75.2} & \textbf{87.2} & \textbf{82.1} & \textbf{88.5} & \textbf{84.8} & \textbf{89.2} & \textbf{85.8 }\\
 \bottomrule
\end{tabular}%
}
\end{table*}
Table~\ref{tab:people} explores the effect of varying the number of tracked subjects while keeping different camera configurations constant. Our approach maintains high LocA and HOTA scores as the number of subjects increases. Even with seven subjects and five cameras, our method achieves a LocA of 89.2\% and a HOTA of 85.8\%, highlighting its robustness in crowded scenes.  While OpenPTrack~\cite{Carraro2019} experiences a decline in accuracy as the number of subjects increases, and Befine~\cite{Boldo2024} struggles with association, COMETH remains reliable across all conditions.

\subsection{Real-World Scenario}\label{SEC:RW-SCENARIO}
\begin{figure*}[t!]
    \centering
    \includegraphics[width=.99\linewidth]{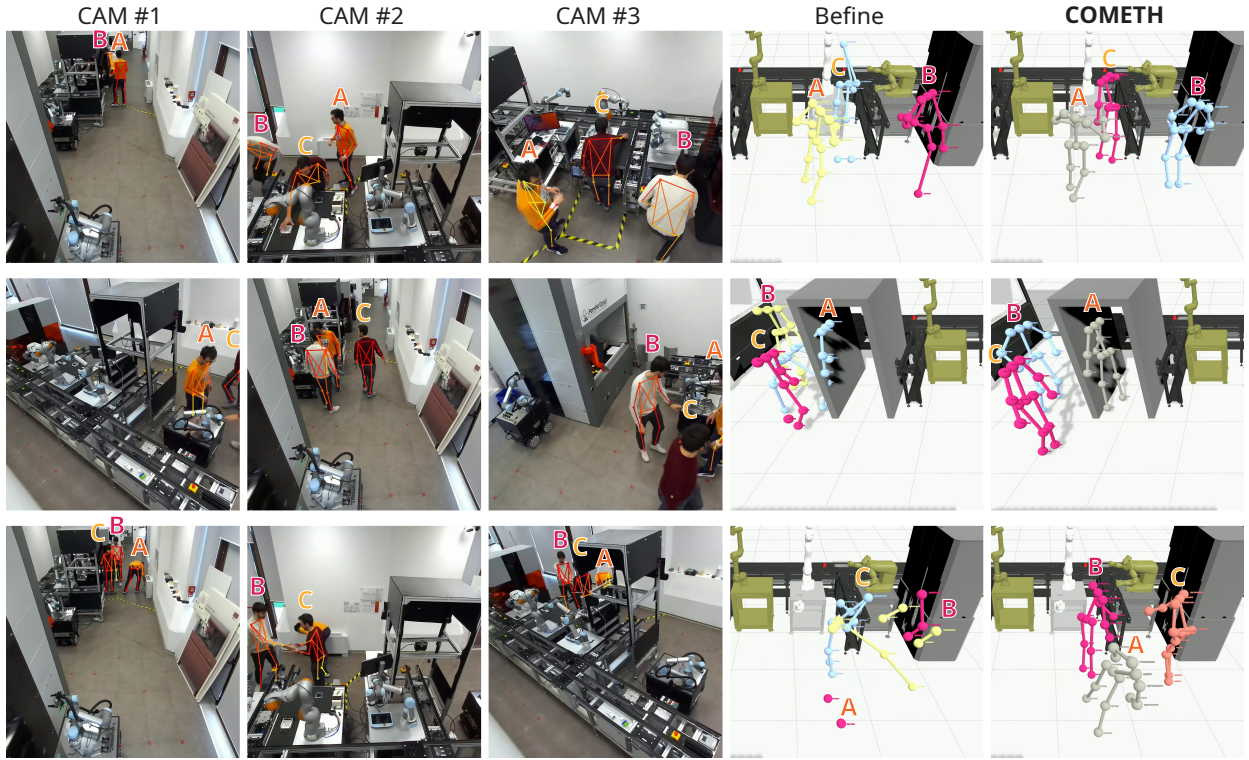}
    \caption{Results on the ICE Laboratory setup from the proposed fusion method. The first three columns are the synchronized frames from the input cameras on which the detections of each HPE are superimposed. The last two columns represent the fused 3D poses resulting respectively from the method proposed by Befine~\cite{Boldo2024} and ours.}
    \label{fig:qualitative}
\end{figure*}


\begin{figure*}[t]
\centering
\begin{subfigure}{0.248\linewidth}
    \centering
    \includegraphics[width=\linewidth]{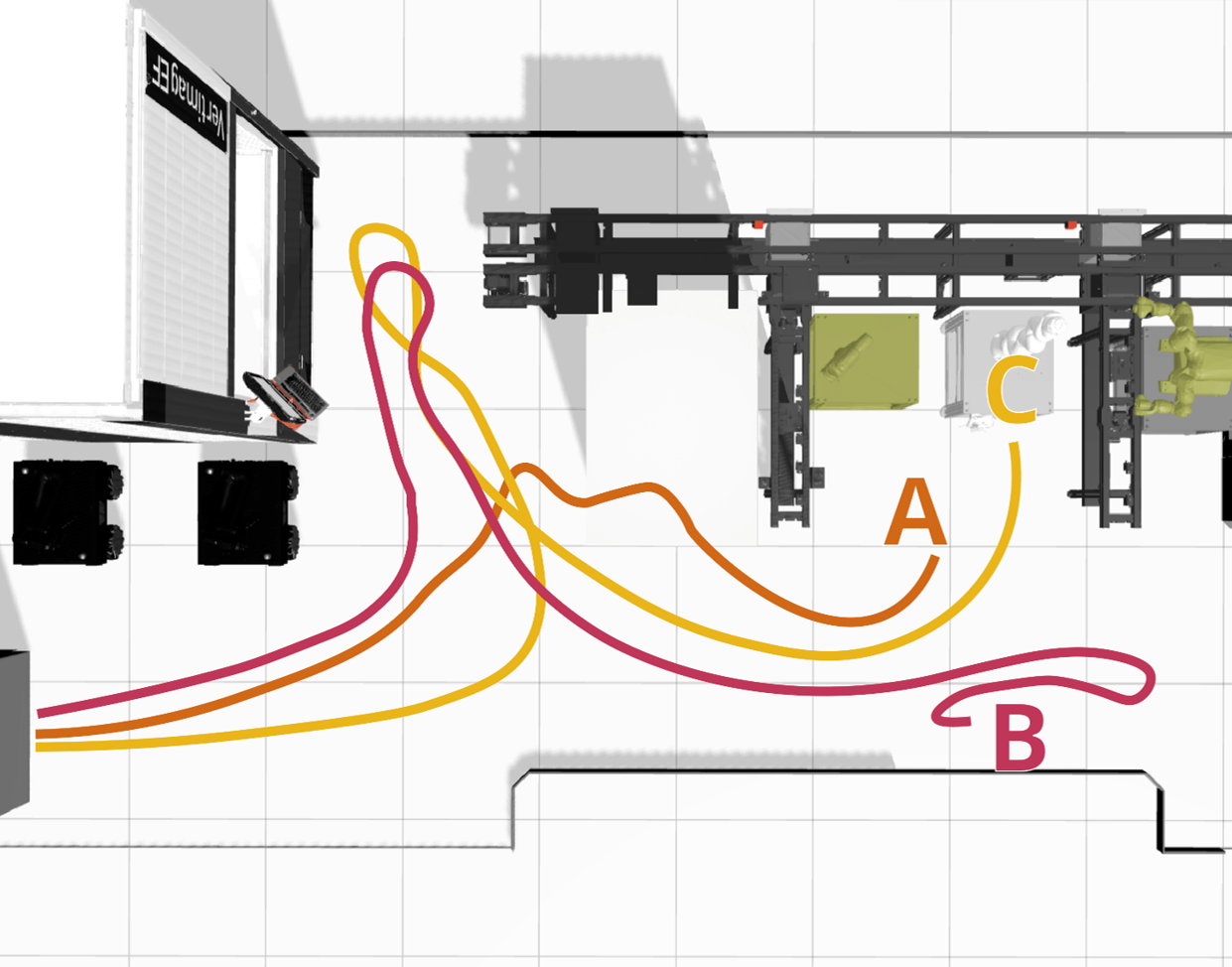}
    \caption{Pseudo-ground truth}
    \label{sf:a}
\end{subfigure}%
\begin{subfigure}{0.248\linewidth}
    \centering
    \includegraphics[width=\linewidth]{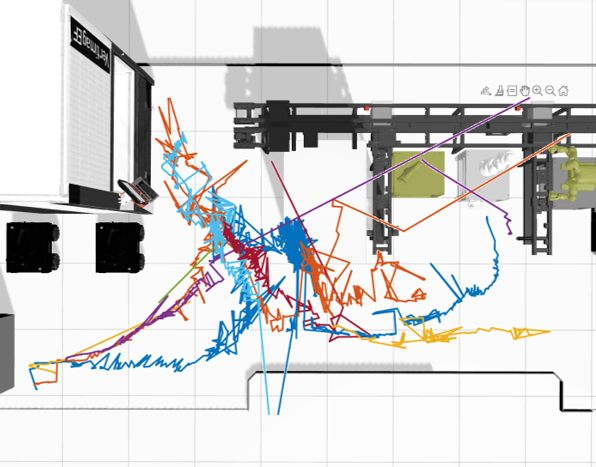}
    \caption{OpenPTrack}
    \label{sf:b}
\end{subfigure}
\begin{subfigure}{0.248\linewidth}
    \centering
    \includegraphics[width=\linewidth]{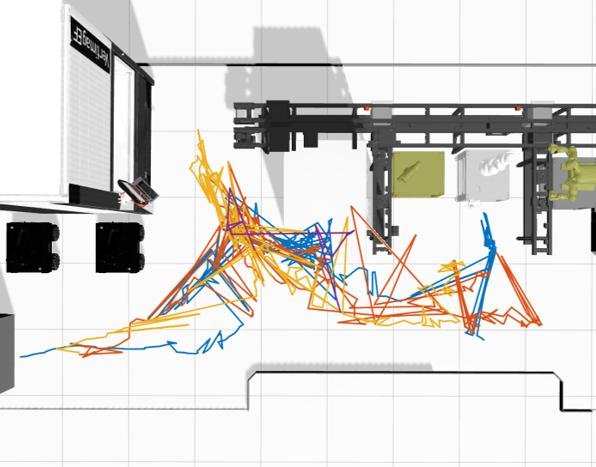}
    \caption{Befine}
    \label{sf:c}
\end{subfigure}%
\begin{subfigure}{0.248\linewidth}
    \centering
    \includegraphics[width=\linewidth]{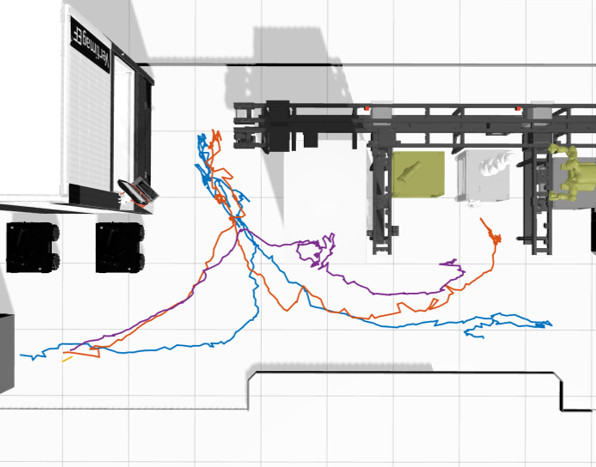}
    \caption{\textbf{COMETH}}
    \label{sf:d}
\end{subfigure}%
\caption{Comparison of the fusion methods for multiple moving subjects on the real-factory scenario, along with the ground truth trajectories. Tracking results are projected on the XY axis of the laboratory. The trajectories of three subjects in \ref{sf:a} are manually annotated from video as reference.} 
\label{fig:projection}
\end{figure*}

Figure~\ref{fig:qualitative} shows the fusion results of the proposed method on three scenes recorded in the ICE Laboratory~\cite{Cunico2024}. We show both Befine~\cite{Boldo2024} and our method's results, where each row represents a separate moment. The detected joints of the HPEs are displayed in a color scale: lighter colors indicate keypoints where the HPE has lower confidence, while darker colors represent those with higher confidence. In the first scene, subject C is fully seen by Cam \#1 and partially seen by Cam \#3. The 3D HPE of Cam \#2 wrongly detects the elbow of C on the surface of the robotic arm. Our aggregator's \textit{outlier removal} block effectively excludes this incorrect detection.
In contrast, Befine~\cite{Boldo2024} fails to correct the keypoint, resulting in an unrealistic configuration. In the second scene, the lower limbs of subject A are only partially seen by Cam \#1. From only the upper body keypoints, our method accurately reconstructs the information about the ankles and knees. In contrast, the clustering algorithm of Befine assigns some keypoints of subject B to subject A. In the third scene, the self-occluding position of subject A poses a significant challenge. Even though the information is noisy and incomplete, and the resulting joint configurations are inaccurate from both aggregators, our method correctly localizes the subject's position in the laboratory.

Figure~\ref{fig:projection} compares the association accuracy of the tracking methods using a top-down perspective of the ICE Laboratory. In \ref{sf:a}, we manually annotated from video the reference trajectories of three subjects (A, B, and C), indicating expected movement paths. The other subfigures represent the results of the fusion methodologies. OpenPTrack~\cite{Carraro2019} exhibits highly fragmented trajectories, with numerous abrupt directional changes and noisy patterns, suggesting significant association errors. Befine~\cite{Boldo2024} also displays inconsistent tracking with a dense concentration of short, jittery trajectories, indicating frequent identity reassignments. Our method demonstrates a more stable and coherent tracking output, where trajectories closely resemble the reference traces, highlighting the robustness of the association algorithm. We uploaded the complete video comparisons as supplementary materials.

\subsection{Performance Analysis}\label{SEC:PERFORMANCE}
\begin{figure}[ht]
    \centering
    \includegraphics[width=.98\linewidth]{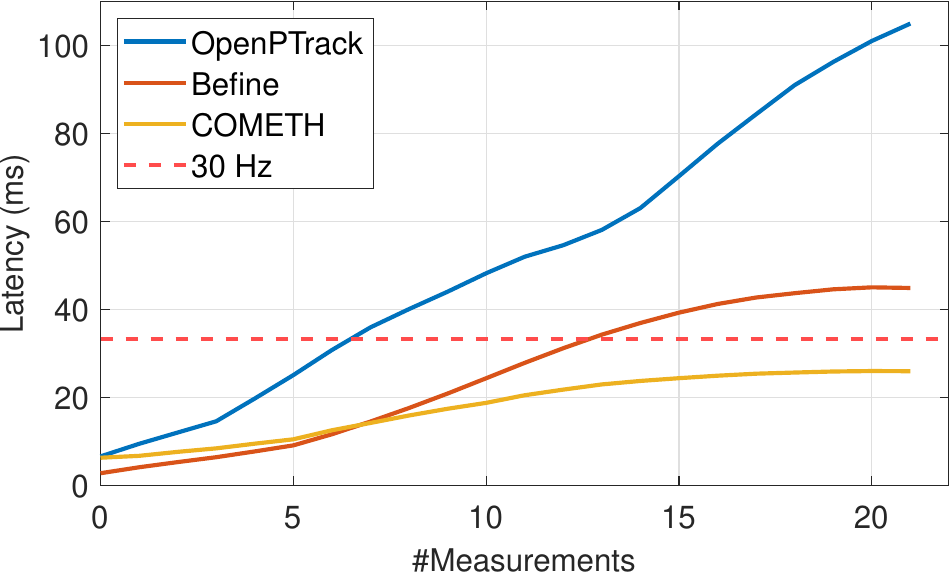}
    \caption{Mean latency of the compared fusion methods when increasing the number of incoming measurements.}
    \label{fig:latency}
\end{figure}
The main purpose of COMETH is to operate in real-time, so our goal was to ensure that the entire pipeline executes in less than 33 ms (i.e., $>$ 30 Hz). Latencies highly depend on the number of incoming measurements, the number of cameras, and the subjects in the scene. Figure~\ref{fig:latency} shows latencies of the compared aggregators when processing data from five edge devices, involving up to seven subjects. While OpenPTrack~\cite{Carraro2019} is linearly dependent on the number of incoming measurements, both Befine~\cite{Boldo2024} and our aggregator's latency do not appear to depend on that.

\section{Conclusion and Future Work} \label{SEC:CONCLUSION}
In this work, we proposed a fast, privacy-aware, multi-view 3D human pose tracking system that effectively integrates multiple camera sources to enhance tracking accuracy. Our approach significantly outperforms state-of-the-art methods in localization, detection, and tracking metrics. The system's ability to robustly track multiple individuals across camera setups highlights its potential for real-world applications.

Future work will explore incorporating additional sensor modalities, such as IMUs, mmWave, or LiDAR, which could improve accuracy under challenging conditions.


\section*{Acknowledgments}
This study was conducted within the MICS (Made in Italy – Circular and Sustainable) Extended Partnership and received funding from Next-Generation EU (Italian PNRR – M4 C2, Invest 1.3 – D.D. 1551.11-10-2022, PE00000004). CUP MICS D43C22003120001 - Cascade funding project CollaborICE. This work has also been supported by the ``PREPARE'' project (n. F/310130/05/X56 - CUP: B39J23001730005) - D.M. MiSE 31/12/2021; and the National Science Foundation (NSF) Foundational Research in Robotics (FRR) program under NSF CAREER Award Grant No. FRR-2443721.


\bibliographystyle{elsarticle-num} 
\bibliography{bibliography}

\begin{thebibliography}{10}
\expandafter\ifx\csname url\endcsname\relax
  \def\url#1{\texttt{#1}}\fi
\expandafter\ifx\csname urlprefix\endcsname\relax\def\urlprefix{URL }\fi
\expandafter\ifx\csname href\endcsname\relax
  \def\href#1#2{#2} \def\path#1{#1}\fi

\bibitem{Lupion2024}
M.~Lupi{\'{o}}n, A.~Polo-Rodr{\'{i}}guez, J.~Medina-Quero, J.~F. Sanjuan, P.~M. Ortigosa, \href{https://doi.org/10.1016/j.inffus.2023.102154}{{3D Human Pose Estimation from multi-view thermal vision sensors}}, Inf. Fusion 104~(November 2023) (2024) 102154.
\newblock \href {https://doi.org/10.1016/j.inffus.2023.102154} {\path{doi:10.1016/j.inffus.2023.102154}}.
\newline\urlprefix\url{https://doi.org/10.1016/j.inffus.2023.102154}

\bibitem{geretti2022process}
L.~Geretti, S.~Centomo, M.~Boldo, E.~Martini, N.~Bombieri, D.~Quaglia, T.~Villa, Process-driven collision prediction in human-robot work environments, in: 2022 IEEE 27th International Conference on Emerging Technologies and Factory Automation (ETFA), IEEE, 2022, pp. 1--8.

\bibitem{rathor2022detailed}
K.~Rathor, S.~Lenka, K.~A. Pandya, B.~Gokulakrishna, S.~S. Ananthan, Z.~T. Khan, A detailed view on industrial safety and health analytics using machine learning hybrid ensemble techniques, in: 2022 International Conference on Edge Computing and Applications (ICECAA), IEEE, 2022, pp. 1166--1169.

\bibitem{Bazo2020}
R.~Bazo, E.~Reis, L.~A. Seewald, V.~F. Rodrigues, C.~A. da~Costa, L.~Gonzaga, R.~S. Antunes, R.~d.~R. Righi, A.~Maier, B.~Eskofier, R.~Fahrig, T.~Horz, {Baptizo: A sensor fusion based model for tracking the identity of human poses}, Inf. Fusion 62~(February) (2020) 1--13.
\newblock \href {https://doi.org/10.1016/j.inffus.2020.03.011} {\path{doi:10.1016/j.inffus.2020.03.011}}.

\bibitem{zhang2022information}
Y.~Zhang, C.~Jiang, B.~Yue, J.~Wan, M.~Guizani, Information fusion for edge intelligence: A survey, Information Fusion 81 (2022) 171--186.

\bibitem{Xu2018}
C.~Xu, J.~He, X.~Zhang, C.~Yao, P.~H. Tseng, {Geometrical kinematic modeling on human motion using method of multi-sensor fusion}, Inf. Fusion 41 (2018) 243--254.
\newblock \href {https://doi.org/10.1016/j.inffus.2017.09.014} {\path{doi:10.1016/j.inffus.2017.09.014}}.

\bibitem{Carraro2019}
M.~Carraro, M.~Munaro, J.~Burke, E.~Menegatti, {Real-time marker-less multi-person 3D pose estimation in RGB-DEPTH camera networks}, Adv. Intell. Syst. Comput. 867 (2019) 534--545.
\newblock \href {http://arxiv.org/abs/1710.06235} {\path{arXiv:1710.06235}}, \href {https://doi.org/10.1007/978-3-030-01370-7_42} {\path{doi:10.1007/978-3-030-01370-7_42}}.

\bibitem{wan2000unscented}
E.~A. Wan, R.~Van Der~Merwe, The unscented kalman filter for nonlinear estimation, in: Proceedings of the IEEE 2000 adaptive systems for signal processing, communications, and control symposium (Cat. No. 00EX373), Ieee, 2000, pp. 153--158.

\bibitem{Cao2019}
Z.~{Cao}, G.~{Hidalgo Martinez}, T.~{Simon}, S.~{Wei}, Y.~A. {Sheikh}, Openpose: Realtime multi-person 2d pose estimation using part affinity fields, IEEE Transactions on Pattern Analysis and Machine Intelligence (2019).

\bibitem{Boldo2024}
M.~Boldo, M.~{De Marchi}, E.~Martini, S.~Aldegheri, D.~Quaglia, F.~Fummi, N.~Bombieri, \href{https://doi.org/10.1016/j.eswa.2024.124089}{{Real-time multi-camera 3D human pose estimation at the edge for industrial applications}}, Expert Syst. Appl. 252~(PA) (2024) 124089.
\newblock \href {https://doi.org/10.1016/j.eswa.2024.124089} {\path{doi:10.1016/j.eswa.2024.124089}}.
\newline\urlprefix\url{https://doi.org/10.1016/j.eswa.2024.124089}

\bibitem{Joo_2017_TPAMI}
H.~Joo, T.~Simon, X.~Li, H.~Liu, L.~Tan, L.~Gui, S.~Banerjee, T.~S. Godisart, B.~Nabbe, I.~Matthews, T.~Kanade, S.~Nobuhara, Y.~Sheikh, Panoptic studio: A massively multiview system for social interaction capture, IEEE Transactions on Pattern Analysis and Machine Intelligence (2017).

\bibitem{vicon}
V.~Nexus, \href{https://documentation.vicon.com/nexus/v2.2/Nexus1\_8Guide.pdf}{Vicon nexus product guide} (2015).
\newline\urlprefix\url{https://documentation.vicon.com/nexus/v2.2/Nexus1\_8Guide.pdf}

\bibitem{desmarais2021review}
Y.~Desmarais, D.~Mottet, P.~Slangen, P.~Montesinos, A review of 3d human pose estimation algorithms for markerless motion capture, Computer Vision and Image Understanding 212 (2021) 103275.

\bibitem{sarafianos20163d}
N.~Sarafianos, B.~Boteanu, B.~Ionescu, I.~A. Kakadiaris, 3d human pose estimation: A review of the literature and analysis of covariates, Computer Vision and Image Understanding 152 (2016) 1--20.

\bibitem{moeslund2006survey}
T.~B. Moeslund, A.~Hilton, V.~Kr{\"u}ger, A survey of advances in vision-based human motion capture and analysis, Computer vision and image understanding 104~(2-3) (2006) 90--126.

\bibitem{zheng2023deep}
C.~Zheng, W.~Wu, C.~Chen, T.~Yang, S.~Zhu, J.~Shen, N.~Kehtarnavaz, M.~Shah, Deep learning-based human pose estimation: A survey, ACM Computing Surveys 56~(1) (2023) 1--37.

\bibitem{martini2023denoising}
E.~Martini, A.~Calanca, N.~Bombieri, Denoising and completion filters for human motion software: a survey with code, Authorea Preprints (2023).

\bibitem{zhang2025multi}
Y.~Zhang, J.~Zhang, S.~Xu, J.~Xiao, Multi-view human pose and shape estimation via mesh-aligned voxel interpolation, Information Fusion 114 (2025) 102651.

\bibitem{denavit1955kinematic}
J.~Denavit, R.~S. Hartenberg, A kinematic notation for lower-pair mechanisms based on matrices (1955).

\bibitem{Koptev2021}
M.~Koptev, N.~Figueroa, A.~Billard, {Real-Time Self-Collision Avoidance in Joint Space for Humanoid Robots}, IEEE Robot. Autom. Lett. 6~(2) (2021) 1240--1247.
\newblock \href {https://doi.org/10.1109/LRA.2021.3057024} {\path{doi:10.1109/LRA.2021.3057024}}.

\bibitem{SMPL:2015}
M.~Loper, N.~Mahmood, J.~Romero, G.~Pons-Moll, M.~J. Black, {SMPL}: A skinned multi-person linear model, ACM Trans. Graphics (Proc. SIGGRAPH Asia) 34~(6) (2015) 248:1--248:16.

\bibitem{delp2007opensim}
S.~L. Delp, F.~C. Anderson, A.~S. Arnold, P.~Loan, A.~Habib, C.~T. John, E.~Guendelman, D.~G. Thelen, Opensim: open-source software to create and analyze dynamic simulations of movement, IEEE transactions on biomedical engineering 54~(11) (2007) 1940--1950.

\bibitem{Werling2023}
K.~Werling, J.~Kaneda, A.~Tan, R.~Agarwal, S.~Skov, T.~Van~Wouwe, S.~Uhlrich, N.~Bianco, C.~Ong, A.~Falisse, S.~Sapkota, A.~Chandra, J.~Carter, E.~Preatoni, B.~Fregly, J.~Hicks, S.~Delp, C.~K. Liu, \href{https://arxiv.org/abs/2406.18537}{Addbiomechanics dataset: Capturing the physics of human motion at scale} (2024).
\newblock \href {https://doi.org/10.48550/ARXIV.2406.18537} {\path{doi:10.48550/ARXIV.2406.18537}}.
\newline\urlprefix\url{https://arxiv.org/abs/2406.18537}

\bibitem{keller2023skel}
M.~Keller, K.~Werling, S.~Shin, S.~Delp, S.~Pujades, C.~K. Liu, M.~J. Black, From skin to skeleton: Towards biomechanically accurate {3D} digital humans, ACM Transaction on Graphics (ToG) 42~(6) (2023) 253:1--253:15.
\newblock \href {https://doi.org/https://doi.org/10.1145/3618381} {\path{doi:https://doi.org/10.1145/3618381}}.

\bibitem{mahmood2019amass}
N.~Mahmood, N.~Ghorbani, N.~F. Troje, G.~Pons-Moll, M.~J. Black, Amass: Archive of motion capture as surface shapes, in: Proceedings of the IEEE/CVF international conference on computer vision, 2019, pp. 5442--5451.

\bibitem{siciliano2008springer}
B.~Siciliano, O.~Khatib, T.~Kr{\"o}ger, Springer handbook of robotics, Vol. 200, Springer, 2008.

\bibitem{boyd2004convex}
S.~P. Boyd, L.~Vandenberghe, Convex optimization, Cambridge university press, 2004.

\bibitem{drillis1964body}
R.~Drillis, R.~Contini, M.~Bluestein, Body segment parameters, Artificial limbs 8~(1) (1964) 44--66.

\bibitem{winter2009biomechanics}
D.~A. Winter, Biomechanics and motor control of human movement, John wiley \& sons, 2009.

\bibitem{martini2024robust}
E.~Martini, H.~Parekh, S.~Peng, N.~Bombieri, N.~Figueroa, A robust filter for marker-less multi-person tracking in human-robot interaction scenarios, in: 2024 33rd IEEE International Conference on Robot and Human Interactive Communication (ROMAN), IEEE, 2024, pp. 424--429.

\bibitem{zwerus2019normative}
E.~L. Zwerus, N.~W. Willigenburg, V.~A. Scholtes, M.~P. Somford, D.~Eygendaal, M.~P. van~den Bekerom, Normative values and affecting factors for the elbow range of motion, Shoulder \& elbow 11~(3) (2019) 215--224.

\bibitem{Ames2019ControlBF}
A.~Ames, S.~D. Coogan, M.~Egerstedt, G.~Notomista, K.~Sreenath, P.~Tabuada, Control barrier functions: Theory and applications, 2019 18th European Control Conference (ECC) (2019) 3420--3431.

\bibitem{levenberg1944method}
K.~Levenberg, A method for the solution of certain non-linear problems in least squares, Quarterly of applied mathematics 2~(2) (1944) 164--168.

\bibitem{kalman1960new}
R.~E. Kalman, A new approach to linear filtering and prediction problems (1960).

\bibitem{Cunico2024}
F.~Cunico, S.~Aldegheri, A.~Avogaro, M.~Boldo, N.~Bombieri, L.~Capogrosso, A.~Caputo, D.~Carra, S.~Centomo, D.~S. Cheng, E.~Cinquetti, M.~Cristani, M.~D. Marchi, F.~Demrozi, M.~Emporio, F.~Fummi, L.~Geretti, S.~Germiniani, A.~Giachetti, F.~Girella, E.~Martini, G.~Menegaz, N.~Muijs, F.~Paci, M.~Panato, G.~Pravadelli, E.~Quintarelli, I.~Siviero, S.~F. Storti, C.~Tadiello, C.~Turetta, T.~Villa, N.~Zannone, D.~Quaglia, Enhancing safety and privacy in industry 4.0: The ice laboratory case study, IEEE Access 12 (2024) 154570--154599.
\newblock \href {https://doi.org/10.1109/ACCESS.2024.3479411} {\path{doi:10.1109/ACCESS.2024.3479411}}.

\bibitem{Luiten2021}
J.~Luiten, A.~Osep, P.~Dendorfer, P.~Torr, A.~Geiger, L.~Leal-Taix{\'{e}}, B.~Leibe, {HOTA: A Higher Order Metric for Evaluating Multi-object Tracking}, Int. J. Comput. Vis. 129~(2) (2021) 548--578.
\newblock \href {http://arxiv.org/abs/2009.07736} {\path{arXiv:2009.07736}}, \href {https://doi.org/10.1007/s11263-020-01375-2} {\path{doi:10.1007/s11263-020-01375-2}}.

\bibitem{kuhn1955hungarian}
H.~W. Kuhn, The hungarian method for the assignment problem, Naval research logistics quarterly 2~(1-2) (1955) 83--97.

\bibitem{Werling2021}
K.~Werling, D.~Omens, J.~Lee, I.~Exarchos, K.~Liu, \href{http://dx.doi.org/10.15607/RSS.2021.XVII.034}{Fast and feature-complete differentiable physics engine for articulated rigid bodies with contact constraints}, in: Robotics: Science and Systems XVII, RSS2021, Robotics: Science and Systems Foundation, 2021.
\newblock \href {https://doi.org/10.15607/rss.2021.xvii.034} {\path{doi:10.15607/rss.2021.xvii.034}}.
\newline\urlprefix\url{http://dx.doi.org/10.15607/RSS.2021.XVII.034}

\bibitem{Lee2018}
J.~Lee, M.~X.~Grey, S.~Ha, T.~Kunz, S.~Jain, Y.~Ye, S.~S.~Srinivasa, M.~Stilman, C.~Karen~Liu, \href{http://dx.doi.org/10.21105/joss.00500}{Dart: Dynamic animation and robotics toolkit}, The Journal of Open Source Software 3~(22) (2018) 500.
\newblock \href {https://doi.org/10.21105/joss.00500} {\path{doi:10.21105/joss.00500}}.
\newline\urlprefix\url{http://dx.doi.org/10.21105/joss.00500}

\bibitem{diamond2016cvxpy}
S.~Diamond, S.~Boyd, Cvxpy: A python-embedded modeling language for convex optimization, Journal of Machine Learning Research 17~(83) (2016) 1--5.

\end{thebibliography}
\end{document}